\documentclass[11pt,a4paper]{article}
\usepackage[hyperref]{emnlp2020}
\usepackage{times}
\usepackage{latexsym}
\usepackage{amssymb}
\usepackage{amsmath}
\usepackage{bbm}

\usepackage{changepage}
\usepackage{graphicx}
\usepackage{subcaption}
\usepackage{float}
\usepackage{cleveref}
\usepackage{booktabs}
\usepackage{makecell}
\usepackage{multirow}
\usepackage{hyperref}
\usepackage{algorithm,algorithmic}

\RequirePackage[textsize=scriptsize]{todonotes}
\RequirePackage{enumitem}
\setlist[itemize]{nosep,leftmargin=*,labelwidth=0pt}
\setlist[enumerate]{nosep}
\setlist[description]{nosep,leftmargin=.8em}

\makeatletter
\g@addto@macro{\normalsize}{%
\setlength{\abovedisplayskip}{0pt}%
\setlength{\abovedisplayshortskip}{0pt}%
\setlength{\belowdisplayskip}{0pt}%
\setlength{\belowdisplayshortskip}{0pt}}
\makeatother

\usepackage{microtype}

\aclfinalcopy 
\def\aclpaperid{3646} 

\newcommand{\shortname}{{\scshape IGL}}
\newcommand{\longname}{Iterative Grid Labeling}
\newcommand{\boldlongname}{\textbf{I}terative \textbf{G}rid \textbf{L}abeling}
\newcommand{\boldshortname}{{\textbf{IGL}}}

\makeatletter
\newcommand\footnoteref[1]{\protected@xdef\@thefnmark{\ref{#1}}\@footnotemark}
\makeatother

\def\ztitle{OpenIE6: Iterative Grid Labeling and Coordination Analysis for Open Information Extraction}

\title{\ztitle}

\newcommand\blfootnote[1]{%
  \begingroup
  \renewcommand\thefootnote{}\footnote{#1}%
  \addtocounter{footnote}{-1}%
  \endgroup
}

\author{
  Keshav Kolluru\textsuperscript{1*}, Vaibhav Adlakha\textsuperscript{1*}, Samarth Aggarwal\textsuperscript{1}, \\  \textbf{Mausam}\textsuperscript{1}, and \textbf{Soumen Chakrabarti}\textsuperscript{2}\\
  \textsuperscript{1} Indian Institute of Technology Delhi\\
  \texttt{keshav.kolluru@gmail.com, vaibhavadlakha95@gmail.com} \\
  \texttt{samarth.aggarwal.2510@gmail.com, mausam@cse.iitd.ac.in} \\
  \textsuperscript{2} Indian Institute of Technology Bombay\\
  \texttt{soumen@cse.iitb.ac.in}
}

\date{}

\begin{document}
\maketitle
\blfootnote{*Equal Contribution}

\begin{abstract}


A recent state-of-the-art neural open information extraction (OpenIE) system generates extractions iteratively, requiring repeated encoding of partial outputs. This comes at a significant computational cost. On the other hand, sequence labeling approaches for OpenIE are much faster, but worse in extraction quality. In this paper, we bridge this trade-off by presenting an iterative labeling-based system that establishes a new state of the art for OpenIE, while extracting 10$\times$ faster. This is achieved through a novel \longname\ (\shortname) architecture, which treats OpenIE as a 2-D grid labeling task. We improve its performance further by applying coverage (soft) constraints on the grid at training time. 

Moreover, on observing that the best OpenIE systems falter at handling coordination structures, our OpenIE system also incorporates a new coordination analyzer built with the same \shortname\ architecture. This \shortname\ based coordination analyzer helps our OpenIE system handle complicated coordination structures, while also establishing a new state of the art on the task of coordination analysis, 
with a 12.3 pts improvement in F1 over previous analyzers.
Our OpenIE system, \textbf{OpenIE6}\footnote{
\url{https://github.com/dair-iitd/openie6}}, beats the previous systems by as much as 4 pts in F1, while being much faster.
\end{abstract}

\section{Introduction}
\label{sec:intro}
Open Information Extraction (OpenIE) is an ontology-free information extraction paradigm that generates extractions of the form \textit{(subject; relation; object)}.  Built on the principles of domain-independence and scalability \cite{mausam16}, OpenIE systems extract open relations and arguments from the sentence, which allow them to be used for a wide variety of downstream tasks like Question Answering \cite{yan&al18, khot&al17}, Event Schema Induction \cite{niranjan&al13} and Fact Salience \cite{marco&al18}.

\begin{figure}[htp]
\centering
\includegraphics[width=8cm,height=2cm]{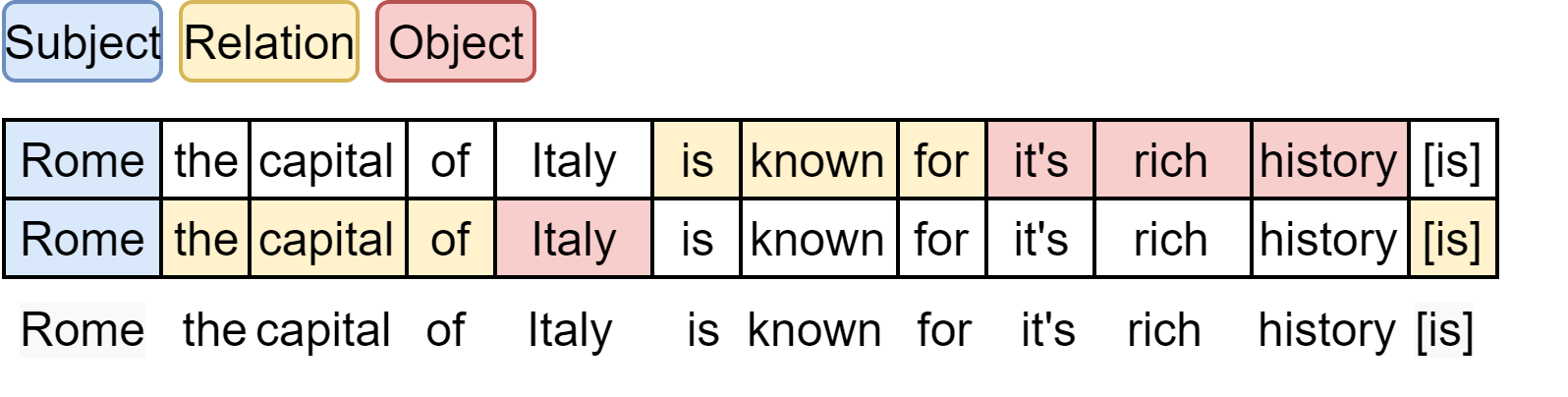}
\vspace*{-4ex}
\caption{The extractions \textit{(Rome; [is] the capital of; Italy)} and \textit{(Rome; is known for; it's rich history)} can be seen as the output of grid labeling. We additionally introduce a token \textit{[is]} to the input.}
\label{fig:grid_example}
\end{figure}
\begin{table*}
\centering
{\small
\begin{tabular}{|l|l|}
\hline
\textbf{Sentence} & \begin{tabular}[c]{@{}l@{}}Other signs of lens subluxation include mild conjunctival redness, vitreous humour degeneration,\\   
 and an increase or decrease of anterior chamber depth .\end{tabular} \\ \hline
\textbf{\shortname} & \begin{tabular}[c]{@{}l@{}}(Other signs of lens subluxation; include; mild conjunctival redness, vitreous humour degeneration)\end{tabular} \\ \hline
\textbf{\begin{tabular}[c]{@{}l@{}}\shortname\\+Constraints\\ \end{tabular}} & \begin{tabular}[c]{@{}l@{}}(Other signs of lens subluxation; include; mild conjunctival redness, vitreous humour degeneration, \\
and an increase or decrease of anterior chamber depth)\end{tabular} \\ \hline
\textbf{\begin{tabular}[c]{@{}l@{}}\shortname\\+Constraints\\+Coordination \\Analyzer\\ \end{tabular}} & \begin{tabular}[c]{@{}l@{}}(Other signs of lens subluxation; include; mild conjunctival redness)\\
(Other signs of lens subluxation; include; vitreous humour degeneration)\\
(Other signs of lens subluxation; include; an increase of anterior chamber depth)\\
(Other signs of lens subluxation; include; an decrease of anterior chamber depth)\end{tabular} \\ \hline
\end{tabular}
}
\caption{For the given sentence, \shortname\ based OpenIE extractor produces an incomplete extraction. Constraints improve the recall by covering the remaining words. Coordination Analyzer handles hierarchical conjunctions.}
\label{tab:redundancy-example}
\end{table*}

End-to-end neural systems for OpenIE have been found to be more accurate compared to their non-neural counterparts, which were built on manually defined rules over linguistic pipelines. The two most popular neural OpenIE paradigms are \emph{generation} \citep{cui+18, kolluru&al20} and \emph{labeling} \citep{stanovsky&al18, roy&al19}. 

\emph{Generation} systems generate extractions one word at a time. IMoJIE \cite{kolluru&al20} is a state-of-the-art OpenIE system that re-encodes the partial set of extractions output thus far when generating the next extraction. This captures dependencies among extractions, reducing the overall redundancy of the output set. However, this repeated re-encoding causes a significant reduction in speed, which limits use at Web scale.


On the other hand, \emph{labeling}-based systems like RnnOIE \cite{stanovsky&al15} are much faster (150 sentences per second, compared to 3 sentences of IMoJIE) but relatively less accurate. They label each word in the sentence as either \textit{S} (Subject), \textit{R} (Relation), \textit{O} (Object) or \textit{N} (None) for each extraction. However, as the extractions are predicted independently, this does not model the inherent dependencies among the extractions.

We bridge this trade-off though our proposed OpenIE system that is both fast and accurate. It consists of an OpenIE extractor based on a novel iterative labeling-based architecture --- \boldlongname\ (\boldshortname). Using this architecture, OpenIE is modeled as a 2-D 
grid labeling problem of size $(M, N)$ where $M$ is a pre-defined maximum number of extractions and $N$ is the sentence length, as shown in Figure~\ref{fig:grid_example}.
Each extraction corresponds to one row in the grid. Iterative assignment of labels in the grid helps \shortname\ capture dependencies among extractions without the need for re-encoding, thus making it much faster than generation-based approaches.

While \shortname\ gives high precision, we can further improve recall by incorporating (soft) global coverage constraints on this 2-D grid. We use constrained training \cite{mehta&al18} by adding a penalty term for all constraint violations. This encourages the model to satisfy these constraints during inference as well, leading to improved extraction quality, without affecting running time.


Furthermore, we observe that existing neural OpenIE models struggle in handling coordination structures, and do not split conjunctive extractions properly.  In response, we first design a new coordination analyzer \cite{ficler&goldberg16b}. It is built with the same \shortname\ architecture, by interpreting each row in the 2-D grid as a coordination structure. This leads to a new state of the art on this task, with a 12.3 pts improvement in F1 over previous best reported result \cite{teranishi+19}, and a 1.8 pts gain in F1 over a strong BERT baseline.


We then combine the output of our coordination analyzer with our OpenIE extractor, resulting in a further increase in performance (Table~\ref{tab:redundancy-example}). 
Our final OpenIE system --- OpenIE6 --- consists of \shortname-based OpenIE extractor (trained with constraints) and \shortname-based coordination analyzer. We evaluate OpenIE6 on four metrics from the literature and find that it exceeds in three of them by at least 4.0 pts in F1. We undertake manual evaluation to reaffirm the gains. In summary, this paper describes OpenIE6, which
\begin{itemize}
    \item is based on our novel \shortname\  architecture,
    \item is trained with constraints to improve recall,
    \item handles conjunctive sentences with our new state-of-art coordination analyzer, which is 12.3 pts better in F1, and
    \item is 10$\times$ faster compared to current state of the art and improves F1 score by as much as 4.0 pts.
\end{itemize}

\section{Related Work}
\label{sec:related}
\citet{banko&al07} introduced the Open Information Extraction  paradigm (OpenIE) and proposed TextRunner, the first model for the task. Following this, many statistical and rule-based systems have been developed  \cite{Fader&al11,etzioni-ijcai11, christensen&al11, Mausam&al12,  corro&al13, angeli&al15, pal&al16, Stanovsky&al2016, saha&al2017,  gashteovski&al17,  saha&mausam18, christina&al18}. 

Recently, supervised neural models have been proposed, which are either trained on extractions bootstrapped from earlier non-neural systems \cite{cui+18}, or on SRL annotations adapted for OpenIE \cite{Stanovsky2016EMNLP}. These systems are primarily of three types, as follows.

\textit{Labeling-based} systems like RnnOIE \cite{stanovsky&al18}, and SenseOIE \cite{roy&al19} identify words that can be syntactic heads of relations, and, for each head word, perform a single labeling to get the extractions.
\citet{jiang&al19} extend these to better calibrate confidences across sentences.
\textit{Generation-based} systems \cite{cui+18,sun2018logician} generate extractions sequentially using seq2seq models. IMoJIE \cite{kolluru&al20}, the current state of art in OpenIE, uses a BERT-based encoder and an iterative decoder that re-encodes the extractions generated so far. This re-encoding captures dependencies between extractions, increasing overall performance, but also makes it 50x slower than RnnOIE.
Recently, \textit{span-based} models \cite{jiang&al19} have been proposed, e.g., SpanOIE \cite{zhan&al19}, which uses a predicate module to first choose potential candidate relation spans, and for each relation span, classifies all possible spans of the sentence as subject or object.

Concurrent to our work \cite{ro&al20} proposed Multi$^2$OIE, a sequence-labeling model for OpenIE, which first predicts all the relation arguments using BERT, and then predicts subject and object arguments associated with each relation using multi-head attention blocks. Their model cannot handle nominal relations and conjunctions in arguments, which can be extracted in our iterative labeling scheme.


\vspace*{0.5ex}
\noindent{\bf OpenIE Evaluation:} Several datasets have been proposed to automatically evaluate OpenIE systems.
OIE2016 \cite{Stanovsky2016EMNLP} introduced an automatically generated reference set of extractions, but it was found to be too noisy with significant missing extractions. Re-OIE2016 \cite{zhan&al19} manually re-annotated the corpus, but did not handle conjunctive sentences adequately. Wire57 \cite{william&al18} contributed high-quality expert annotations, but for a small corpus of 57 sentences. We use the CaRB dataset \cite{bhardwaj&al19}, which re-annotated OIE2016 corpus via crowd-sourcing. 

The benchmarks also differ in their scoring functions along two dimensions: (1)~computing similarity for each (gold, system) extraction pair, (2)~defining a mapping between system and gold extractions using this similarity. OIE16 computes similarity by serializing the arguments into a sentence and finding the number of matching words. It maps each system extraction to one gold (one-to-one mapping) to compute both precision and recall. Wire57 uses the same one-to-one mapping but computes similarity at an argument level. CaRB uses one-to-one mapping for precision but maps multiple gold to the same system extraction (many-to-one mapping) for recall. Like Wire57, CaRB computes similarity at an argument level.


\vspace{0.5ex}
\noindent \textbf{OpenIE for Conjunctive Sentences}: Performance of OpenIE systems can be further improved by identifying coordinating structures governed by conjunctions (e.g., `and'),  and splitting conjunctive extractions (see Table \ref{tab:redundancy-example}). We follow CalmIE \cite{saha&mausam18}, which is part of OpenIE5 system -- it splits a conjunctive sentence into smaller sentences based on detected coordination boundaries, and runs OpenIE on these split sentences to increase overall recall. 



For detecting coordination boundaries, \citet{ficler&goldberg16a} re-annotate the Penn Tree Bank corpus with coordination-specific tags. Neural parsers trained on this data use similarity and replacability of conjuncts as features \cite{ficler&goldberg16b, teranishi+17}. The current state-of-the-art system \citep{teranishi+19} independently detects coordinator, begin, and end of conjuncts, and does joint inference using Cocke–Younger–Kasami (CYK) parsing over context-free grammar (CFG) rules. Our end-to-end model obtains better accuracy than this approach.



\begin{figure}[htp]
\centering
\includegraphics[height=3.5cm, width=8cm]{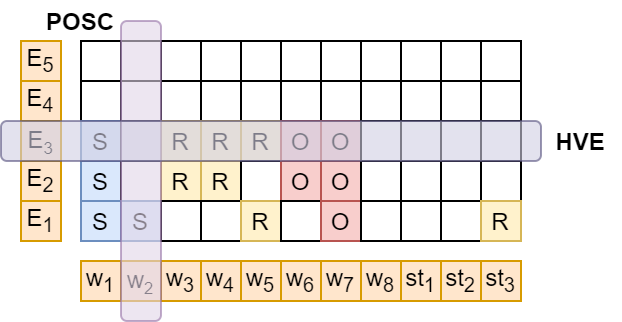}
\hfill
\vspace*{-2ex}
\caption{2-D grid for OpenIE with extraction as rows and words as columns. The values represent the labels (S)ubject, (R)elation, (O)bject. The empty cells represent (N)one. Constraints are applied across rows (\textbf{HVE}) and columns (\textbf{POSC}).}
\label{fig:oie_grid}
\end{figure}

\vspace{0.5ex}
\noindent{\bf Constrained Training:} 
Constraining outputs of the model is a way to inject prior knowledge into deep neural networks \cite{zhiting&al16, xu&al18, nandwani&al19}. These constraints can be applied either during training or inference or both. We follow \citet{mehta&al18}, which models an output constraint as a differentiable penalty term defined over output probabilities given by the network. This penalty is combined with the original loss function for better training.

\citet{bhutani&al19} propose an OpenIE system to get extractions from question-answer pairs.
Their decoder enforces vocabulary and structural constraints on the output both during training and inference.
In contrast, our system uses constraints only during training.



\section{Iterative Grid Labeling for OpenIE}
\label{sec:il}
Given a sentence with word tokens $\{w_1, w_2, \ldots, w_N\}$ the task of OpenIE is to output a set of extractions, say $\{E_1, E_2, \ldots, E_M\}$, where each extraction is of the form \textit{(subject; relation; object)}. For a labeling-based system, each word is labeled as \textit{S} (Subject), \textit{R} (Relation), \textit{O} (Object), or \textit{N} (None) for every extraction. We model this as a  2-D grid labeling problem of size $(M,N)$, where the words represent the columns and the extractions represent the rows (Figure~\ref{fig:oie_grid}). The output at position $(m,n)$ in the grid ($L_{m,n}$) represents the label assigned to the $n^{th}$ word in the $m^{th}$ extraction. 

We propose a novel \boldlongname\ (\shortname) approach to label this grid, filling up one row after another iteratively. We refer to the OpenIE extractor trained using this approach as \shortname-OIE.

\begin{figure}[t]
\centering
\includegraphics[height=6cm, width=8cm]{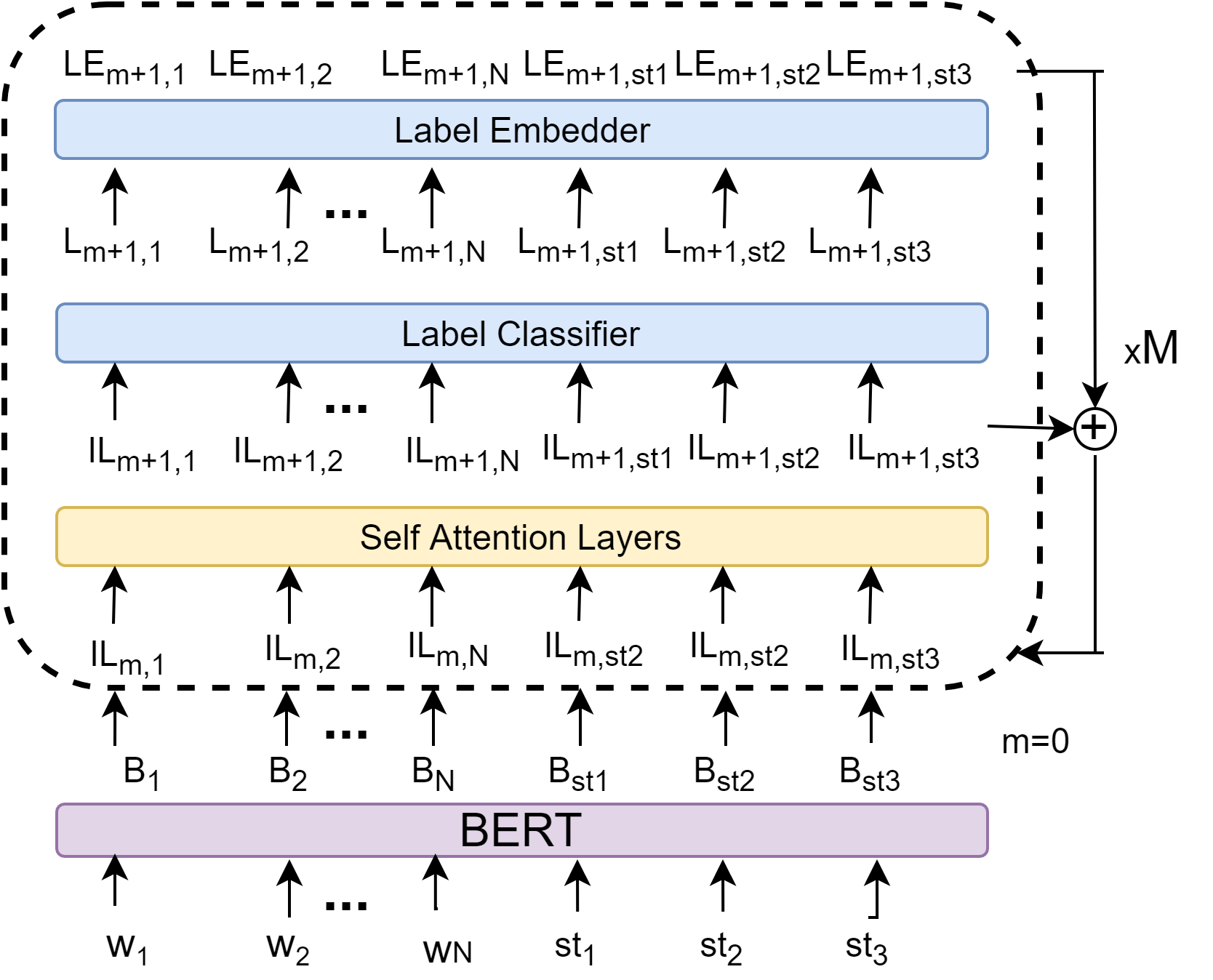}
\hfill
\vspace*{-3ex}
\caption{Model architecture for \shortname. BERT-embeddings of the words are iteratively passed through self-attention layers. $st_1$, $st_2$, $st_3$ refer to the appended tokens \emph{[is]}, \emph{[of]}, \emph{[from]}, respectively. At every iteration, we get an extraction by labeling the words using a fully-connected layer. Embeddings of the generated labels are added to the iterative layer embeddings before passing them to the next iteration.} 
\label{fig:model_architecture}
\end{figure}

\shortname-OIE is based on a BERT encoder, which computes contextualized embeddings for each word. The input to the BERT encoder is $\{w_1,$ $w_2, \ldots, w_N, \emph{[is]}, \emph{[of]}, \emph{[from]}\}$. The last three tokens (referred as $st_i$ in Figure~\ref{fig:model_architecture}) are appended because, sometimes, OpenIE is required to predict tokens that are not present in the input sentence.\footnote{`is', `of' and `from' are the most frequent such tokens.}  E.g., ``\textit{US president Donald Trump gave a speech on Wednesday.}'' will have one of the extractions as \textit{(Donald Trump; [is] president [of]; US)}. The appended tokens make such extractions possible in a labeling framework.

The contextualized embeddings for each word or appended token are iteratively passed through a 2-layer transformer to get their \textit{IL embeddings} at different levels, until a maximum level $M$, i.e. a word $w_n$ has a different contextual embedding $IL_{m,n}$ for every row (level) $m$. At every level $m$, each $IL_{m,n}$ is passed though a fully-connected labeling layer to get the labels for words at that level (Figure~\ref{fig:model_architecture}). Embeddings of the predicted labels are added to the \textit{IL embeddings} before passing them to the next iteration. This, in principle, maintains the information of the extractions output so far, and hence can capture dependencies among labels of different extractions.
For words that were broken into word-pieces by BERT, only the embedding of the first word-piece is retained for label prediction. We sum the  cross-entropy loss between the predicted labels and the gold labels at every level to get the final loss, denoted by $J_{CE}$.

OpenIE systems typically assign a confidence value to an extraction. In \shortname, at every level, the respective extraction is assigned a confidence value by adding the log probabilities of 
the predicted labels (\textit{S}, \textit{R}, and \textit{O}), and normalizing this by the extraction length.

We believe that \shortname\ architecture has value beyond OpenIE, and can be helpful in tasks where a set of labelings for a sentence is desired, especially when labelings have dependencies amongst them.\footnote{\shortname\ is a generalization of \citet{ju&a18}. Their model can only label spans which are subsets of one another.}  We showcase another application of \shortname~for the task of coordination analysis in Section \ref{sec:conjunctions}.

\begin{figure*}[htp]
\includegraphics[width=0.98\textwidth]{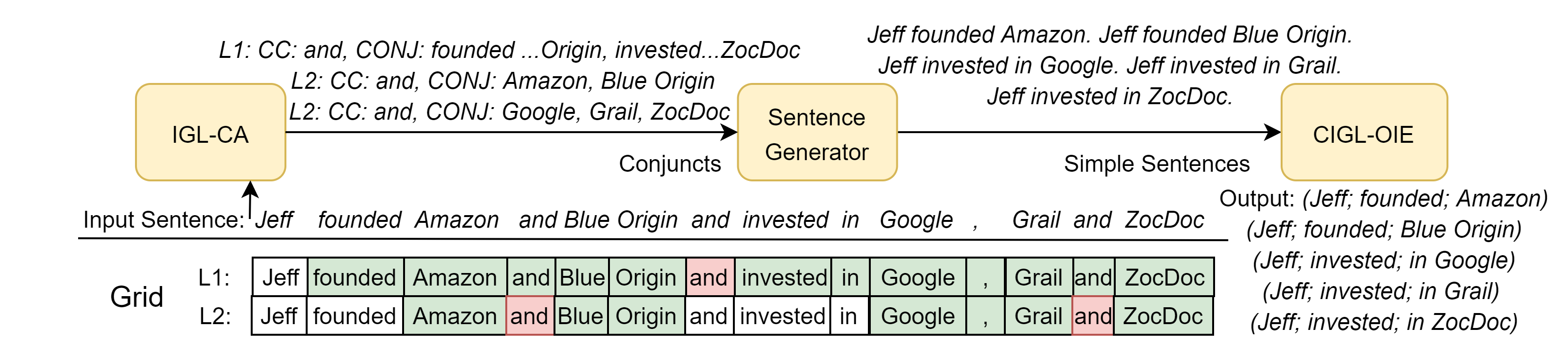}
\hfill
\vspace*{-1ex}
\caption{The final OpenIE system. \shortname-CA identifies conjunct boundaries by labeling a 2-D grid. This generates simple sentences and C\shortname-OIE emits the final extractions.}
\label{fig:oie_system}
\end{figure*}

\section{Grid Constraints}
\label{sec:constraints}
Our preliminary experiments revealed that \shortname-OIE has good precision, but misses out important extractions. In particular, we observed that the set of output extractions did not capture all the information from the sentence (Table~\ref{tab:redundancy-example}). We formulate constraints over the 2-D grid of extractions (as shown in Figure~\ref{fig:oie_grid}) which act as an additional form of supervision to improve the coverage. We implement these as soft constraints, by imposing additional violation penalties in the loss function.
This biases the model to learn to satisfy the constraints, without explicitly enforcing them at inference time.


To describe the constraints, we first define the notion of a \emph{head verb} as all verbs except light verbs (do, be, is, has, etc.). We run a POS tagger on the input sentence, and find all head verbs in the sentence by removing all light verbs.\footnote{We used the light verbs listed by \citet{jain&al16}.}  For example, for the sentence, \textit{``Obama gained popularity after Oprah endorsed him for the presidency"}, the head verbs are \textit{gained} and \textit{endorsed}. In order to cover all 
valid
extractions like \textit{(Obama; gained; popularity)} and \textit{(Oprah; endorsed him for; the presidency)}, we design the following coverage constraints: 


\begin{itemize}
\item \textit{POS Coverage} (\textbf{POSC}): All words with POS tags as nouns (N), verbs (V), adjectives (JJ), and adverbs (RB) should be part of at least one extraction. 
E.g. the words \textit{Obama}, \textit{gained}, \textit{popularity}, \textit{Oprah}, \textit{endorsed}, \textit{presidency} must be covered in the set of extractions.

\item \textit{Head Verb Coverage} (\textbf{HVC}): Each head verb 
should be present in the relation span of some (but not too many) extractions.
E.g. \textit{(Obama; gained; popularity)}, \textit{(Obama; gained; presidency)} is not a comprehensive set of extractions.

\item \textit{Head Verb Exclusivity} (\textbf{HVE}): The relation span of one extraction can contain at most one head verb.
E.g. \textit{gained popularity after Oprah endorsed} is not a good relation as it contains two head verbs.

\item \textit{Extraction Count} (\textbf{EC}): The total number of extractions with head verbs in the relation span must be no fewer than the number of head verbs in the sentence.
In the example, there must be at least two extractions containing head verbs, as the sentence itself has two head verbs.
\end{itemize}

\noindent\textbf{Notation}: 
We now describe the penalty terms for these constraints. Let $p_n$ be the POS tag of $w_n$. We define an indicator $x^{imp}_n=1$ if $p_n\in \{\text{N, V, JJ, RB}\}$, and $0$ otherwise. Similarly, let $x_n^{hv}=1$ denote that $w_n$ is a head verb. At each extraction level $m$, the model computes $Y_{mn}(k)$, the probability of assigning the $n^{th}$ word the label $k \in \{\text{S, R, O, N}\}$.  We formulate the penalties associated with our constraints as follows:

\begin{itemize} 
    \item \textbf{POSC} - To ensure that the $n^{th}$ word is covered, we compute its maximum probability ($posc_n$) of belonging to any extraction. We introduce a penalty if this value is low. This penalty is aggregated over words with important POS tags, $J_{posc} = \sum_{n=1}^{N}x^{imp}_n \cdot posc_n$, where
    \begin{equation*} \label{constraint_pos}
    \begin{aligned}
        posc_n = 1 - \max_{m \in [1,M]}\left( \max_{k \in \{\text{S,R,O}\}}Y_{mn}(k)\right) \\
    \end{aligned}
    \end{equation*}

    \item \textbf{HVC} - A penalty is imposed for the $n^{th}$ word, if it is not present in relation of any extraction 
    or if it is present in relation of many extractions.
    This penalty is aggregated over head verbs, $J_{hvc} = \sum_{n=1}^{N}x^{hv}_n \cdot hvc_n$, where 
    $hvc_n = \left| 1 - \sum_{m=1}^M Y_{mn}(R)\right|$.
    
    \item \textbf{HVE} - A penalty is imposed if the relation span of an extraction contains more than one head verb. This penalty is summed over all extractions. I.e., $J_{hve} = \sum_{m=1}^{M} hve_{m}$, where
    \begin{equation*}
    \begin{aligned}
        hve_m = \max\left(0, \left(\sum\limits_{n=1}^{N} x^{hv}_n \cdot Y_{mn}(R) \right) - 1 \right) \\
    \end{aligned}
    \end{equation*}
    \item \textbf{EC} - $ec_m$ denotes the score $\in [0,1]$ of the $m^{th}$ extraction containing a head verb, i.e. $ec_m = \max_{n \in [1,N]}\left( x^{hv}_n \cdot Y_{mn}(R)\right)$. A penalty is imposed if the sum of these scores is less than the actual number of head verbs in the sentence. 
    \begin{equation*}
    \begin{aligned}
        J_{ec} = \max\left(0, \sum_{n=1}^{N}x^{hv}_n - \sum_{m=1}^{M}ec_m\right)
    \end{aligned}
    \end{equation*}
\end{itemize}

Ideally, no constraint violations of \textbf{HVC} and \textbf{HVE} would imply that \textbf{EC} would also never gets violated. However, as these are soft constraints, this scenario is never materialized in practice. We find that our model performs better and results in fewer constraint violations when trained with \textbf{POSC}, \textbf{HVC}, \textbf{HVE} and \textbf{EC} combined. The full loss function is $J = J_{CE}+\lambda_{posc}  J_{posc}+\lambda_{hvc} J_{hvc}+\lambda_{hve} J_{hve}+\lambda_{ec} J_{ec}$, where $\lambda_\star$ are hyperparameters. We refer to the OpenIE extractor trained using this constrained loss as Constrained \longname\ OpenIE Extractor (C\shortname-OIE).

The model is initially trained without constraints for a fixed \textit{warmup} number of iterations, followed by constrained training till convergence.

\section{Coordination Boundary Detection}
\label{sec:conjunctions}
Coordinated conjunctions (CC) are conjunctions such as \textit{``and"}, \textit{``or"} that connect, or coordinate words, phrases, or clauses (they  are called the conjuncts). The goal of coordination analysis is to detect  coordination structures --- the coordinating conjunctions along with their constituent conjuncts. In this section we build a novel coordination analyzer and use its output downstream for OpenIE.

Sentences can have hierarchical coordinations, i.e., some coordination structures nested within the conjunct span of others \cite{saha&mausam18}.
Therefore, we pose coordination analysis as a hierarchical labeling problem, as illustrated in Figure~\ref{fig:oie_system}. We formulate a 2-D grid labeling problem, where all coordination structures at the same hierarchical level are predicted in the same row.

Specifically, we define a grid of size $(M, N)$, where $M$ is the maximum depth of hierarchy and $N$ is the number of words in the sentence. The value at $(m,n)^{th}$ position in the grid represents the label assigned to the $n^{th}$ word in the $m^{th}$ hierarchical level, which can be \textit{CC} (coordinated conjunction), \textit{CONJ} (belonging to a conjunct span), or \textit{N} (None). 
Using \shortname\ architecture for this grid gives an end-to-end Coordination Analyzer that can detect multiple coordination structures, with two or more conjuncts. We refer to this Coordination Analyzer as \shortname-CA.  

\vspace{0.5ex}
\noindent{\bf Coordination Analyzer in OpenIE:}
Conjuncts in a coordinate structure exhibit \emph{replaceability} -- a sentence is still coherent and consistent, if we replace a coordination structure with any of its conjuncts \cite{ficler&goldberg16b}. Following CalmIE's approach, we generate simple (non-conjunctive) sentences using \shortname-CA. We then run C\shortname-OIE on these simple sentences to generate extractions. These extractions are de-duplicated and merged to yield the final extraction set (Figure~\ref{fig:oie_system}).
This pipelined approach describes our final OpenIE system --- \textbf{OpenIE6}.



For a conjunctive sentence, C\shortname-OIE's confidence values for extractions will be with respect to multiple simple sentences, and may not be calibrated across them. We use a separate confidence estimator, consisting of a BERT encoder and an LSTM decoder trained on (sentence, extraction) pairs. It computes a log-likelihood for every extraction w.r.t. the original sentence --- this serves as a better confidence measure for OpenIE6.

\section{Experimental Setup}
\label{sec:datasets_exp}
\begin{table*}
\centering {\footnotesize
\begin{tabular}{lcccccccrcc} 
\toprule       
System & \multicolumn{2}{c}{CaRB} & \multicolumn{2}{c}{CaRB(1-1)} & \multicolumn{2}{c}{OIE16-C} & \multicolumn{1}{c}{Wire57-C} & \multicolumn{1}{c}{Speed}
        \\ \cmidrule(r){2-3} \cmidrule(r){4-5} \cmidrule(r){6-7} \cmidrule(r){8-8} \cmidrule(r){9-9}
      & F1 & AUC & F1 & AUC & F1 & AUC & F1 & \makecell{Sentences/sec.}     \\
       
\midrule                         
MinIE           
& 41.9 & - & 38.4 & - & 52.3 & - & 28.5 & 8.9\\
ClausIE         
& 45.0 & 22.0 & 40.2 & 17.7 & 61.0 & 38.0 & 33.2 & 4.0\\
OpenIE4            
& 51.6 & 29.5 & 40.5 & 20.1 & 54.3 & 37.1 & 34.4 & 20.1\\
OpenIE5            
& 48.0 & 25.0 & 42.7 & 20.6 & 59.9 & 39.9 & 35.4 & 3.1\\
\midrule
SenseOIE        
& 28.2 & -    & 23.9 & -    & 31.1 & -    & 10.7 & - \\
SpanOIE         
& 48.5 & -    & 37.9 & -    & 54.0 & -    & 31.9 & 19.4\\
RnnOIE          
& 49.0 & 26.0 & 39.5 & 18.3 & 56.0 & 32.0 & 26.4 & \textbf{149.2}\\
\cite{cui+18}       
& 51.6 & 32.8 & 38.7 & 19.8 & 53.5 & 37.0 & 33.3 & 11.5\\
IMoJIE
& 53.5 & 33.3 & 41.4 & 22.2 & 56.8 & 39.6 & 36.0 & 2.6\\
\shortname-OIE &
52.4 & 33.7 & 41.1 & 22.9 & 55.0 & 36.0 & 34.9 & 142.0 \\
C\shortname-OIE & 
\textbf{54.0} & \textbf{35.7} & 42.8 & 24.6 & 59.2 & 40.0 & 36.8 & 142.0 \\
C\shortname-OIE + \shortname-CA (OpenIE6) &
52.7 & 33.7 & \textbf{46.4} & \textbf{26.8} & \textbf{65.6} & \textbf{48.4} & \textbf{40.0} & 31.7 \\

\bottomrule
\end{tabular}
\caption{Evaluation of OpenIE. Using constrained learning, C\shortname-OIE gives better scores on all metrics compared to IMoJIE. Adding a coordination analyzer, C\shortname-OIE + \shortname-CA (OpenIE6) gives the best scores in 3 of the 4 metrics. MinIE, SenseOIE, SpanOIE do not output confidences. Code of SenseOIE is not available to compute speed. 
}
\label{tab:main-table}
}
\end{table*}

We train OpenIE6 using the OpenIE4 training dataset used to train IMoJIE\footnote{Available from \href{https://github.com/dair-iitd/imojie}{github:dair-iitd/imojie}}. It has 190,661 extractions from 92,774 Wikipedia sentences.
We convert each extraction to a sequence of labels over the sentence. This is done by looking for an exact string match of the words in the extraction with the sentence. In case there are multiple string matches for one of the arguments of the extraction, we choose the string match closest to the other arguments. This simple heuristic covers almost 95\% of the training data. We ignore the remaining extractions that have multiple string matches for more than one argument.

We implement our models using Pytorch Lightning \cite{falcon&19}. We 
use pre-trained weights of ``BERT-base-cased''\footnote{\label{footnote:huggingface}\href{https://github.com/huggingface/transformers}{github:huggingface/transformers}} for OpenIE extractor and ``BERT-large-cased''\footnoteref{footnote:huggingface} for coordination analysis.
We do not use BERT-large for OpenIE extractor as we observe almost same performance with a significant increase in computational costs.
We set the maximum number of iterations, $M$=5 for OpenIE and $M$=3 for Coordination Analysis. We use the SpaCy POS tagger\footnote{\href{https://spacy.io}{https://spacy.io}} for enforcing constraints. The various hyper-parameters used are mentioned in Appendix~\ref{sec:reproducibility}.

\vspace{0.5ex}
\noindent
{\bf Comparison Systems:}  We compare OpenIE6 against several recent neural and non-neural systems. These include generation (IMoJIE and \citet{cui+18}\footnote{We use the BERT implementation available at \href{https://github.com/dair-iitd/imojie}{github:dair-iitd/imojie}}), labeling (RnnOIE, SenseOIE) and span-based (SpanOIE) systems.
We also compare against non-neural baselines of MinIE \cite{gashteovski&al17},
ClausIE \cite{corro&al13},
OpenIE4 \cite{christensen&al11}\footnote{\href{https://github.com/allenai/openie-standalone}{github:allenai/openie-standalone}} and OpenIE5 \cite{saha&al2017, saha&mausam18}.\footnote{\href{https://github.com/dair-iitd/OpenIE-standalone}{github:dair-iitd/openie-standalone}}  We use open-source implementations for all systems except SenseOIE, for which the code is not available and we use the system output provided by the authors.

\vspace{0.5ex}
\noindent
{\bf Evaluation Dataset and Metrics:} We evaluate all systems against CaRB's reference extractions, as they have higher coverage and quality compared to other datasets. Apart from CaRB's scoring function, we also use scoring functions of OIE16 and Wire57 benchmarks on the CaRB reference set, which we refer to as \textit{OIE16-C} and \textit{Wire57-C}. Additionally we use \textit{CaRB(1-1)}, a variant of CaRB that retains CaRB's similarity computation, but uses a one-to-one mapping for both precision and recall (similar to OIE16-C,  Wire57-C). 

For each system, we report a final F1 score using precision and recall computed by these scoring functions. OpenIE systems typically
associate a confidence value with each extraction, which can be varied to generate a precision-recall (P-R) curve. We also report the area under P-R curve (AUC) for all scoring functions except Wire57-C, as its matching algorithm is not naturally compatible with P-R curves.
We discuss details of these four metrics in 
Appendix~\ref{sec:metrics}.

For determining the speed of a system, we analyze the number of sentences it can process per second. We run all the systems on a common set of 3,200 sentences \cite{stanovsky&al18}, using a V100 GPU  and 4 cores of Intel Xeon CPU (the non-neural systems use only the CPU).

\begin{table}
\centering {\footnotesize
\begin{tabular}{lccccccc}
\toprule
System                    
& Precision & Yield & \makecell{Total\\Extrs}\\
\midrule       
C\shortname-OIE
& 77.9 & 131 & 174 \\
OpenIE6
& \textbf{78.8} & \textbf{222} & \textbf{291} \\
\bottomrule
\end{tabular}}
\caption{Manual comparison of Precision and Yield on 100 random conjunctive sentences from CaRB Gold.}
\label{tab:manual_annotation}
\end{table}
\vspace*{-1ex}

\section{Experiments and Results}
\label{sec:experiments}
\begin{table*}
\centering {\footnotesize
\begin{tabular}{lcccccccccccc}
\toprule
System & \multicolumn{1}{c}{Wire57-C} & \multicolumn{2}{c}{CaRB} & \multicolumn{5}{c}{Constraint Violations} &  \multicolumn{1}{c}{\multirow{2}{*}{\makecell{Num. of\\Extrs}}}\\
    \cmidrule(r){2-2} \cmidrule(r){3-4} \cmidrule(r){5-9}
    & F1 & F1 & AUC & POSC & HVC & HVE & EC & HVC+HVE+EC & \\
\midrule                          
IMoJIE
& 36.0 & 53.5 & 33.3 & 687 & 521 & 105 & 330 & 957 & 1354 \\
\shortname-OIE
& 34.9 & 52.4 & 33.7 & 1494 & 375 & \textbf{128} & 284 & 787 & 1401\\
\shortname-OIE (POSC)
& 36.7 & 49.6 & 33.4 & \textbf{396} & 303 & 200 & \textbf{243} & 746 & \textbf{1577}\\
\shortname-OIE (HVC,HVE,EC)
& 35.8 & 53.2 & 32.7 & 1170 & 295 & 144 & 246 & \textbf{655} & 1509\\
C\shortname-OIE
& \textbf{36.8} & \textbf{54.0} & \textbf{35.7} & 766 & \textbf{274} & 157 & 237 & 668 & 1531\\
\midrule
Gold
& 100 & 100 & 100 & 371 & 324 & 272 & 224 & 820 & 2714 \\
\bottomrule
\end{tabular}}
\caption{Performance and number of constraint violations for training with different sets of constraints. C\shortname-OIE represents training \shortname\ architecture based OpenIE extractor with all the constraints - POSC, HVC, HVE and EC}
\label{tab:const_ablation}
\end{table*}
\subsection{Speed and Performance}
\textit{How does OpenIE6 compare in speed and performance?}


Table \ref{tab:main-table} reports the speed and performance comparisons across all metrics for OpenIE.  We find that the base OpenIE extractor --- \shortname-OIE --- achieves a 60$\times$ speed-up compared to IMoJIE, while being lower in performance by 1.1 F1, and better in AUC by 0.4 pts, when using CaRB scoring function.

We find that training \shortname-OIE along with constraints (C\shortname-OIE), helps to improve the performance without affecting inference time. 
This system is better than all previous systems over all the considered metrics. It beats IMoJIE by (0.5, 2.4) in CaRB (F1, AUC) and 0.8 F1 in Wire57-C. 

Further, adding the coordination analyzer module (\shortname-CA) gives us OpenIE6, which is 10$\times$ faster than IMoJIE (32 sentences/sec) and achieves significant improvements in performance in 3 of the 4 metrics considered. It improves upon IMoJIE in F1 by 5.0, 8.8, 4.0 pts in CaRB(1-1), OIE16-C and Wire57-C, respectively. However, in the CaRB metric, adding this module leads to a decrease of (1.5, 0.9) pts in (F1, AUC).

On closer analysis, we notice that the current scoring functions for OpenIE evaluation do not handle conjunctions properly. CaRB over-penalizes OpenIE systems for incorrect coordination splits whereas other scoring functions under-penalize them. This is also evidenced in the lower CaRB scores of for both OpenIE-5\footnote{OpenIE5 uses CalmIE for conjunctive sentences.} (vs. OpenIE4) and
OpenIE6
(vs. C\shortname-OIE) --- the two systems that focus on conjunctive sentences. We trace this issue to the difference in mapping used for recall computation (one-to-one vs many-to-one). We refer the reader to Appendix~\ref{appendix:conjunctive} for a detailed analysis of this issue.

To resolve this variation in different scoring functions, we undertake a manual evaluation. Two annotators (authors of the paper), blind to the underlying systems (C\shortname-OIE and
OpenIE6),
independently label each extraction as correct or incorrect for a subset of 100 conjunctive sentences. Their inter-annotator agreement is 93.46\% (See Appendix~\ref{sec:manual_comparison} for details of manual annotation setup). After resolving the extractions where they differ, we report the precision and yield in Table \ref{tab:manual_annotation}. Here, yield is the number of correct extractions generated by a system. It is a surrogate for recall, since its denominator, number of all correct extractions, is hard to annotate for OpenIE.  

We find that
OpenIE6
significantly increases the yield (1.7$\times$) compared to C\shortname-OIE along with a marginal increase in precision. This result underscores the importance of splitting coordination structures for 
OpenIE. 

\subsection{Constraints Ablation}

\textit{How are constraint violations related to model performance?}

We divide the constraints into two groups: one which is dependent on head verb(s): \{HVC, HVE and EC\}, and the other which is not -- POSC. We separately train \shortname\ architecture based OpenIE extractor with these two groups of constraints, and compare them with no constraints (\shortname-OIE), all constraints (C\shortname-OIE) and IMoJIE. In Table \ref{tab:const_ablation}, we report the performance on Wire57-C and CaRB, and also report the number of constraint violations in each scenario.

Training \shortname\ architecture based OpenIE extractor with POSC constraint (\shortname-OIE (POSC)), leads to a reduction in POSC violations. However, the number of violations of (HVC+HVE+EC) remains high. On the other hand, training only with head verb constraints (HVC,HVE,EC) reduces their violations but the POSC violations remains high. Hence, we find that training with all the constraints achieves the best performance. Compared to \shortname-OIE, it reduces the POSC violation from 1494 to 766 and (HVC+HVE+EC) violations from 787 to 668. The higher violations of Gold may be attributed to an overall larger number of extractions in the reference set.

\subsection{Coordination Analysis}
\textit{How does our coordination analyzer compare against other analyzers? How much does the coordination analyzer benefit OpenIE systems?}

Following previous works \cite{teranishi+17, teranishi+19}, we evaluate two variants of our \shortname\ architecture based coordination analyzer (\shortname-CA) -- using BERT-Base and BERT-Large, on coordination-annotated Penn Tree Bank \cite{ficler&goldberg16a}. We compute the Precision, Recall and F1 of the predicted conjunct spans. In Table~\ref{tab:coord_analysis}, we find that both BERT-Base and BERT-Large variants outperform the previous state-of-art \cite{teranishi+19} by 9.4 and 12.3 F1 points respectively. For fair comparison, we train a stronger variant of \citet{teranishi+19}, replacing the LSTM encoder with BERT-Base and BERT-Large. Even in these settings, \shortname-CA performs better by 1.8 and 1.3 F1 points respectively, highlighting the significance of our \shortname\ architecture. Overall, \shortname-CA
establishes a new state of the art for this task.

\begin{table}
\centering {\footnotesize
\begin{tabular}{lccccccc}
\toprule
System                    
& \multicolumn{1}{c}{Precision} & \multicolumn{1}{c}{Recall} & \multicolumn{1}{c}{F1}\\
\midrule                          
\cite{teranishi+17}
& 71.5 & 70.7 & 71.0 \\
\cite{teranishi+19}
& 75.3 & 75.6 & 75.5 \\
\midrule
BERT-Base:
& & \\
\cite{teranishi+19}
& 83.1 & 83.2 & 83.1 \\
\shortname-CA
& 86.3 & 83.6 & 84.9 \\
\midrule
BERT-Large:
& & \\
\cite{teranishi+19}
& 86.4 & 86.6 & 86.5 \\
\shortname-CA
& \textbf{88.1} & \textbf{87.4} & \textbf{87.8} \\     
\bottomrule
\end{tabular}}
\caption{P, R, F1 of the system evaluated on Penn Tree Bank for different systems. We use both BERT-Base and BERT-Large as the encoder}
\label{tab:coord_analysis}
\end{table}

\begin{table}
\centering {\footnotesize
\begin{tabular}{lccc}
\toprule
Coordination Analyzer & IMoJIE & C\shortname-OIE \\ \hline
None
& 36.0 & 36.8 \\
CalmIE
& 37.7 & 38.0 \\
\cite{teranishi+19}
& 36.1 & 36.5 \\
\shortname-CA
& \textbf{39.5} & \textbf{40.0} \\  
\bottomrule
\end{tabular}}
\caption{Wire57 F1 scores of IMoJIE and C\shortname-OIE with addition of different coordination analyzers. \shortname-CA improves both of the OpenIE extractors.}
\label{tab:oie_parsers}
\end{table}

To affirm that the gains of better coordination analysis help the downstream OpenIE task, we experiment with using different coordination analyzers with C\shortname-OIE and IMoJIE.
From Table~\ref{tab:oie_parsers}, we see a considerable improvement in the downstream OpenIE task using \shortname-CA for both IMoJIE and C\shortname-OIE, which we attribute to better conjunct-boundary detection capabilities of the model. For C\shortname-OIE, this gives a 2 pts increase in Wire57-C F1, compared to CalmIE's coordination analyzer (CalmIE-CA).


\section{Error Analysis}
\label{sec:error_analysis}

We examine extractions from a random sample of 50 sentences from CaRB validation set, as output by OpenIE6. We identify three major sources of errors in these sentences: \newline
\textbf{Grammatical errors:} (24\%) We find that the sentence formed by serializing the extraction is not grammatically correct. We believe that combining our extractor with a pre-trained language model might help reduce such errors. \newline
\textbf{Noun-based relations:} (16\%) These involve introducing additional words in the relation span. Although our model can introduce \textit{[is], [of], [from]} in relations (Section~\ref{sec:il}), it may miss some words for which it was not trained. E.g. \textit{[in]} in \textit{(First Security; based [in]; Salt Lake City)} for the phrase \textit{Salt Lake City-based First Security}. \newline
\textbf{Lack of Context:} (10\%) Neural models for OpenIE including ours, do not output extraction context \cite{Mausam&al12}. E.g. for  ``\textit{She believes aliens will destroy the Earth}'', the extraction \textit{(Context(She believes); aliens; will destroy; the Earth)} can be misinterpreted without the context.

We also observe incorrect boundary identification for relation argument (13\%),  cases in which coordination structure in conjunctive sentences are incorrectly split (11\%), lack of coverage (4\%) and other miscellaneous errors (18\%).

\section{Conclusion}
\label{sec:conclusion}
We propose a new OpenIE system -- OpenIE6, based on the novel \longname\ architecture, which models sequence labeling tasks with overlapping spans as a 2-D grid labeling problem. OpenIE6 is 10x faster, handles conjunctive sentences and establishes a new state of art for OpenIE. We highlight the role of constraints in training for OpenIE. Using the same architecture, we achieve a new state of the art for coordination parsing, with a 12.3 pts improvement in F1 over previous analyzers. We plan to explore the utility of this architecture in other NLP problems. OpenIE6 is available at \url{https://github.com/dair-iitd/openie6} for further research.

\section*{Acknowledgements}
We thank the anonymous reviewers for their suggestions and feedback. Mausam is supported by IBM AI Horizons Network grant, an IBM SUR award, grants by Google, Bloomberg and 1MG, Jai Gupta Chair Fellowship and Visvesvaraya faculty award by Govt. of India. We thank IIT Delhi HPC facility for compute resources. Soumen was partly supported by a Jagadish Bose Fellowship and an AI Horizons Network grant from IBM.


%

\bibliography{anthology,emnlp2020}
\bibliographystyle{acl_natbib}

\clearpage

\appendix

\section{Metrics}
\label{sec:metrics}
\begin{table*}
\centering {\footnotesize
\begin{tabular}{lccccccc}
\toprule

& System 1 (P, R, F1) & System 2 (P, R, F1)\\
\midrule       
\makecell{Talks resumed between USA and China\\Gold: \\(Talks; resumed; between USA and China)\\}
& \makecell{(Talks; resumed; between USA)\\(Talks; resumed; between China)\\CaRB: (50.0, 66.7, 57.1)\\ CaRB (1-1): (50.0, 66.7, 57.1)} & \makecell{(Talks; resumed; between USA and China)\\ \\ CaRB: (100, 100, 100)\\ CaRB (1-1): (100, 100, 100)} \\
\midrule
\makecell{I ate an apple and orange\\Gold: \\(I; ate; an apple) \\ (I; ate; an orange)}
& \makecell{(I; ate; an apple)\\(I; ate; an orange)\\ CaRB: (100, 100, 100)\\ CaRB (1-1): (100, 100, 100)} & \makecell{(I; ate; an apple and an orange)\\ \\ CaRB: (57.1, 100, \textbf{72.7}) \\ CaRB (1-1): (53.5, 50.0, \textbf{57.1})}\\
\bottomrule
\end{tabular}}
\caption{Evaluation of CaRB and CaRB (1-1) on two sentences.}
\label{tab:metrics}
\end{table*}

\subsection{Introduction}

Designing an evaluation benchmark for an under-specified and subjective task like OpenIE has gathered much attention. Several benchmarks, consisting of gold labels and scoring functions have been contributed. While coverage and quality of gold labels of these benchmarks have been extensively studied, differences in their scoring functions is largely unexplored.
We evaluate all our systems on the CaRB reference set, which has 641 sentences and corresponding human annotated extractions in both dev and test set. As the underlying gold labels, is the same, system performances differ only due to difference in design choices of these scoring functions, which we explore in detail here.

\subsection{Scoring Functions of Benchmarks} \label{appendix:scoring_functions}

\textbf{OIE2016}\footnote{\url{https://github.com/gabrielStanovsky/oie-benchmark}} creates a one-to-one mapping between (gold, system) pairs by serializing the extractions and comparing the number of common words within them. Hence the system is not penalized for misidentifying parts of an one argument in another.
Precision and recall for the system are computed using the one-to-one mapping obtained, i.e. precision is (no. of system extractions mapped to gold extractions)/ (total no. of system extractions) and recall is (no. of gold extractions mapped to system extractions)/(total no. of gold extractions). These design choices have several implications \cite{william&al18, bhardwaj&al19}. Overlong system extractions which are mapped, are not penalized, and extractions with partial coverage of gold extractions, which are not mapped, are not rewarded at all.

\noindent \textbf{Wire57}\footnote{\url{https://github.com/rali-udem/WiRe57}} attempts to tackle the shortcomings of OIE2016. For each gold extraction, a set of candidate system extractions are chosen on the basis of whether they share at least one word for each of the arguments\footnote{We refer to \textit{subject}, \textit{relation} and \textit{object} as \textit{arguments} of the extraction.} of the extraction, with the gold. It then creates a one-to-one mapping by greedily matching gold with one of the candidate system extraction on the basis of token-level F1 score. Token level precision and recall of the matches are then aggregated to get the score for the system. Computing scores at token level helps in penalizing overly long extractions. 

Wire57 ignores the confidence of extraction and reports just the F1 score (F1 at zero confidence). One way to generate AUC for Wire57 is by obtaining precision and recall scores at various confidence levels by passing a subset of extractions to the scorer. However, due to Wire57's criteria of matching extractions on the basis of F1 score, the recall of the system does not decrease monotonically with increasing confidence, which is a requirement for calculating AUC.

OIE2016 and Wire57 both use one-to-one mapping strategy, due to which a system extraction, that contains information from multiple gold extractions, is unfairly penalized.

\noindent \textbf{CaRB}\footnote{\url{https://github.com/dair-iitd/CaRB}} also computes similarity at a token level, but it is slightly more lenient than Wire57 --- it considers number of common words in (gold,system) pair for each argument of the extraction. However, it uses one-to-one mapping for precision and
many-to-one mapping for computing recall. While this solves the issue of penalizing extractions with information from multiple gold extractions, it inadvertently creates another one --- unsatisfactorily evaluating systems which split on conjunctive sentences. We explore this in detail in the next section.

\subsection{CaRB on Conjunctive Sentences}\label{appendix:conjunctive}

Coordinate structure in conjunctive sentences are of two types: 
\begin{itemize}
    \item \textit{Combinatory}, where splitting the sentence by replacing the coordinate structure with one of the conjuncts can lead to incoherent extractions. E.g. splitting ``\textit{Talks resumed between USA and China}'' will give \textit{(Talks; resumed; between USA)}.
    \item \textit{Segregatory}, where splitting on coordinate structure can lead to shorter and coherent extractions. E.g. splitting ``\textit{I ate an apple and orange.}'' gives \textit{(I; ate; an apple)} and \textit{(I; ate; an orange)}.
\end{itemize}

\noindent Combinatory coordinate structures are hard to detect (in some cases even for humans). Some systems (ClausIE, CalmIE and ours) use some heuristics such as not splitting if coordinate structure is preceded by ``\textit{between}''. In all other cases, coordinate structure is treated as segregatory, and is split.

The human-annotated gold labels of CaRB dataset correctly handle conjunctive sentences in most of the cases. However, we find that compared to scoring function of OIE2016 and Wire57, CaRB over-penalizes systems for incorrectly splitting combinatory coordinate structures.

We trace this issue to the difference in mapping used for recall computation (one-to-one vs many-to-one). 

Consider two systems -- System 1, which splits on all conjunctive sentences (without any heuristics), and System 2, which does not. For the sentence ``\textit{I ate an apple and orange}'', the set of gold extractions are \{\textit{(I; ate; an apple)}, \textit{(I; ate; orange)}\}. System 2, which (incorrectly) doe not split on the coordinate structure, gets a perfect recall score of 1.0, similar to System 1, which correctly splits the extractions (Table~\ref{tab:metrics}). On the other hand, when System 2 incorrectly splits extractions for the sentence ``\textit{Talks resumed between USA and China}'', it is penalized on both precision and recall by CaRB, giving it a much lower score than System 2.

Due to this phenomena, we find that the gains obtained by our system on splitting the segregatory coordinate structures correctly is overshadowed by penalties of incorrectly splitting the coordinate structures. To re-affirm this, we evaluate all the systems on \textbf{CaRB(1-1)}, a variant of CaRB which retains all the properties of CaRB, except that it uses one-to-one mapping for computing recall.

We notice that our C\shortname-OIE+\shortname-CA shows improvements in CaRB(1-1) and other metrics which use one-to-one mapping (OIE16, Wire57) (Table~\ref{tab:main-table}). But it shows a decrease in CaRB score. This demonstrates that the primary reason for the decrease in performance is the many-to-one mapping in CaRB.

However, we also observe that this is not the best strategy for evaluation as it assigns equal score to both the cases --- splitting a combinatory coordinate structure, and not splitting a segregatory coordinate structure (Table~\ref{tab:metrics}). This is also not desirable as a long extraction which is not split is better than two incorrectly split extractions. Hence, we consider that one-to-one mapping for computing recall under-penalizes splitting a combinatory coordinate structure.

Determining the right penalty in this case is an open-ended problem. We leave it to further research to design an optimal metric for evaluating conjunctive sentences for OpenIE.

\section{Reproducibility}
\label{sec:reproducibility}

\paragraph{Compute Infrastructure:} We train all of our models using a Tesla V100 GPU (32 GB). 
\paragraph{Hyper-parameter search:}
The final hyper-parameters used during train our model are listed in \tablename~\ref{tab:hyperparams}. We also list the search-space, which was manually tuned. We select the model based on the best CaRB (F1) score on validation set. 
\paragraph{Validation Scores:} We report the best validation scores in \tablename~\ref{tab:val-table}.
\paragraph{Number of parameters:} The C\shortname-OIE model contains 110 million parameters and \shortname-CA contains 335 million parameters. The difference is because they use BERT-base and BERT-large models, respectively.

\begin{table*}
\centering
\begin{tabular}{lccccccc}
\toprule
Hyperparameters & Best Values & Grid Search \\
\midrule   
Training: \\
Batch Size & 24 & \{16,32,24\}\\
Optimizer & AdamW & \{AdamW, Adam\}\\
Learning Rate & $2\times10^{-5}$ & \{$1\times10^{-3}$, $2\times10^{-4}$, $5\times10^{-5}$\}\\
\midrule
Model: \\
Iterative Layers & 2 & \{1,2,3\} \\
$\lambda_{posc}$ & 3 & \{0.1, 1, 3, 5, 10\} \\
$\lambda_{hvc}$ & 3 & \{0.1, 1, 3, 5, 10\} \\
$\lambda_{hve}$ & 3 & \{0.1, 1, 3, 5, 10\} \\
$\lambda_{ec}$ & 3 & \{0.1, 1, 3, 5, 10\} \\
\bottomrule
\end{tabular}
\caption{Hyperparameter settings.}
\label{tab:hyperparams}
\end{table*}

\begin{table*}

\centering {\footnotesize
\begin{tabular}{lccccccccccc} 
\toprule       
System & \multicolumn{2}{c}{CaRB} & \multicolumn{2}{c}{CaRB(1-1)} & \multicolumn{2}{c}{OIE16-C} & \multicolumn{1}{c}{Wire57-C} 
        \\ \cmidrule(r){2-3} \cmidrule(r){4-5} \cmidrule(r){6-7} \cmidrule(r){8-8}
       & F1 & AUC & F1 & AUC & F1 & AUC & F1 \\
\midrule      
IMoJIE & 
55.2 & 35.2 & 43.1 & 23.4 & 59.0 & 42.5 & 38.7 \\
\shortname-OIE & 
53.4 & 32.7 & 41.8 & 22.0 & 56.8 & 36.6 & 36.9 \\
C\shortname-OIE & 
\textbf{55.2} & \textbf{35.5} & 43.9 & 23.9 & 62.3 & 42.4 & 39.1 \\
C\shortname-OIE + \shortname-CA (OpenIE6) &
53.8 & 35.0 & \textbf{47.5} & \textbf{27.7} & \textbf{67.7} & \textbf{51.9} & \textbf{42.4} \\
\bottomrule
\end{tabular}
\caption{Evaluation of OpenIE systems on validation set}
\label{tab:val-table}
}
\end{table*}

\section{Manual Comparison}
\label{sec:manual_comparison}

The set of extractions from both the systems, C\shortname-OIE and OpenIE6 were considered for a random 100 conjunctive sentences from the validation set. We identify a conjunctive sentence, based on the predicted conjuncts of coordination analyzer. The annotators are instructed to check if the extraction has well formed arguments and is implied by the sentence. 

\begin{figure*}[htp]
\includegraphics[width=0.99\textwidth]{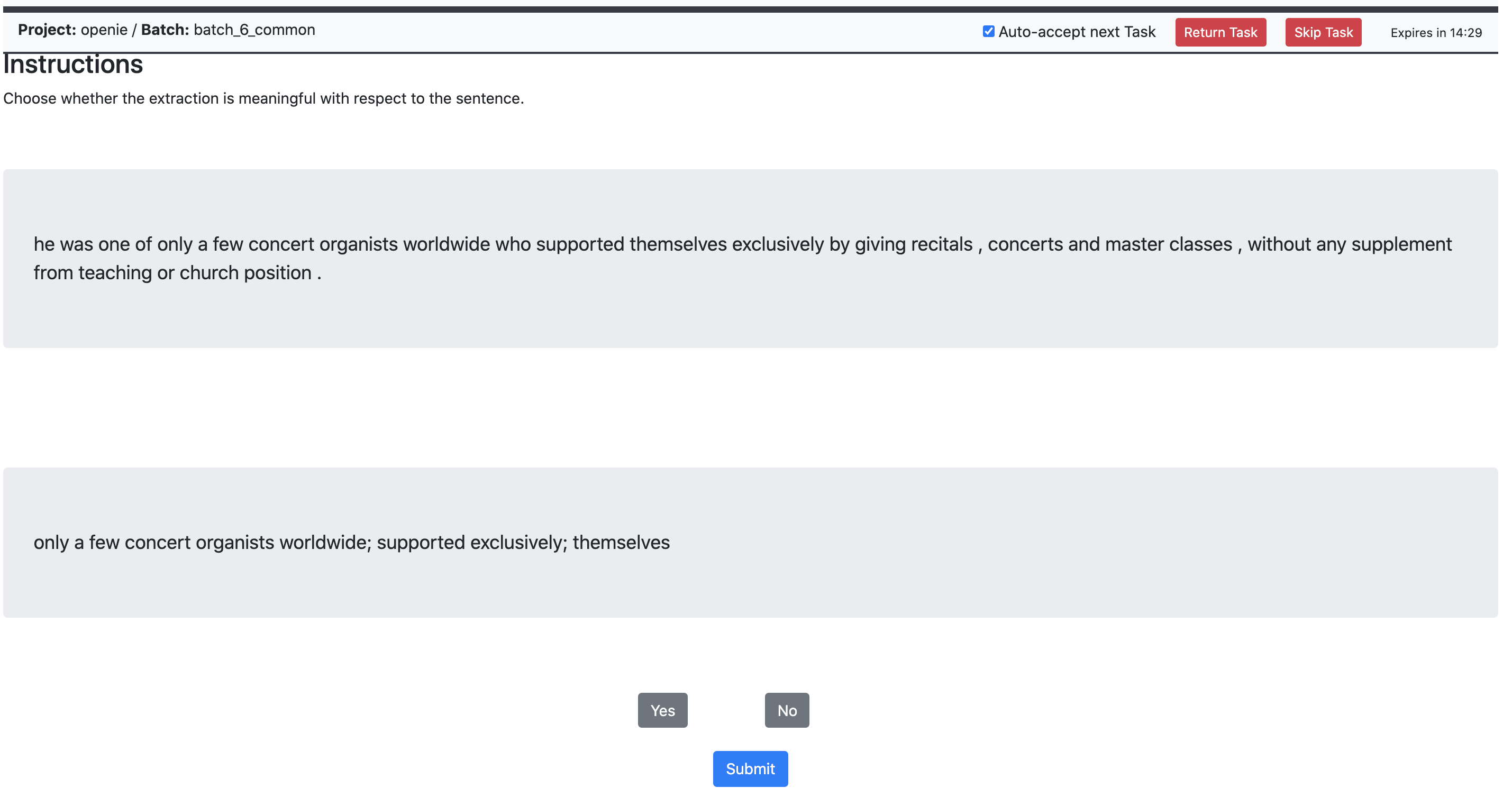}
\caption{Process for manual comparison. Each extraction from both the systems are presented to the annotator in a randomized order. The annotator checks if the extraction can be inferred from the original sentence and marks it accordingly.}
\label{fig:manual_comparison}
\end{figure*}

A screenshot of the process is shown in Figure~\ref{fig:manual_comparison}.

\end{document}